\documentclass[]{spie}  

\usepackage{amsmath,amsfonts,amssymb}
\usepackage[colorlinks=true, allcolors=blue]{hyperref}

\usepackage{cite}
\usepackage{amsmath,amssymb,amsfonts}
\usepackage{algorithmic}
\usepackage{url}
\usepackage{textcomp}
\usepackage{xcolor}
\usepackage{color}
\usepackage[pdftex]{graphicx}
\def\BibTeX{{\rm B\kern-.05em{\sc i\kern-.025em b}\kern-.08em
T\kern-.1667em\lower.7ex\hbox{E}\kern-.125emX}}
\usepackage[hang,small,bf]{caption}
\usepackage[subrefformat=parens]{subcaption}
\newcommand{\ocut}[1]{}

\newcommand{\mycut}[1]{}

\newcommand{\etextd}[1]{}
\newcommand{\jtextd}[1]{}

\title{Depth estimation from 4D light field videos}

\author[a]{Takahiro Kinoshita}
\author[a]{Satoshi Ono}
\affil[a]{Kagoshima University, Japan}

\authorinfo{E-mail: \{sc115015,~ono\}@ibe.kagoshima-u.ac.jp}

\begin{document} 
\maketitle

\begin{abstract}
\jtextd{
   4Dライトフィールド（LF）画像を用いた奥行き（視差）推定は，
   ここ数年，活発に研究されている．従来研究では，静的な4D LF画像
   を用いた奥行き推定に焦点が当てられており，時間情報が利用されておらず，
   LF映像を用いて奥行きを推定する研究は行われていない． 本論文では，
   4D LF映像を用いた奥行き推定を目的としたエンドツーエンドの学習を
   可能にするニューラルネットワークアーキテクチャを提案する．
   また，深層学習モデルを十分に訓練可能な中規模4D LF映像データセットを構築する．
   合成映像および実撮影映像の双方を用いた実験により，
   時間情報を活用することでノイズが多い領域の奥行き推定が可能であることを示した．
}
Depth (disparity) estimation from 4D Light Field (LF) images
has been a research topic for the last couple of years.
Most studies have focused on depth estimation from static 4D LF images
while not considering temporal information, i.e., LF videos.
This paper proposes an end-to-end neural network architecture 
for depth estimation from 4D LF videos.
This study also constructs a medium-scale synthetic 4D LF video dataset that 
can be used for training deep learning-based methods.
Experimental results using synthetic and real-world 4D LF videos 
show that temporal information contributes to the improvement of 
depth estimation accuracy in noisy regions.
Dataset and code is available at: 
\url{https://mediaeng-lfv.github.io/LFV_Disparity_Estimation}.
\end{abstract}



\section{INTRODUCTION}
\label{sec:intro}  
\jtextd{
   コンピュテーショナルフォトグラフィ（Computational Photography: CP）は，
   撮影後の画像処理により通常のカメラでは撮影できなかった写真を合成できるため，
   近年大きな注目を集めている．
   主要なCP技術の1つであるライトフィールドイメージングは，
   異なる複数の方向からの光線量を捉えることができる．
   2つの空間パラメータと2つの角度パラメータで表現される
   4Dライトフィールド画像（Light Field Images: LFIs）の
   撮影は，カメラアレイや
   1台の移動カメラで構成されたガントリー，
   LytroやRaytrixのような特別に設計された
   プレノプティックカメラ~\cite{ng2005light}等により行える．
   4D LFIsに記録された空間角度情報は，
   デジタルリフォーカス~\cite{ng2005light}，
   素材分類~\cite{wang20164d}，
   超解像~\cite{zhang2019residual}，
   最近では奥行き推定~\cite{tao2013depth,shin2018epinet,leistner2019learning}
   など
   多くのコンピュータビジョンアプリケーションの性能向上に利用されている．
}

Computational photography has attracted much attention
because post-image processing after shooting can synthesize photos
that cannot be taken with a normal camera.
Light field imaging%
, one of the major
computational photography techniques,  captures the amount of light
rays from several different directions.
4D light field images (LFIs), which are represented by two spatial and two angular parameters, 
can be captured by camera arrays, 
gantries consisting of a single moving camera, and 
specially designed plenoptic cameras~\cite{ng2005light} such as Lytro and Raytrix.
The spatial-angular information recorded in 4D LFIs
has been exploited to improve the performance of many computer
vision applications, such as digital refocusing~\cite{ng2005light},
image segmentation,
material classification,
super-resolution,
and, most recently, depth estimation
~\cite{tao2013depth,shin2018epinet,faluvegi20193d}.

\begin{figure}[tbp]
  \vspace{-1.8em}
   \centering
   \begin{minipage}{0.49\hsize}
      \centering
      \includegraphics[height=4.5cm]{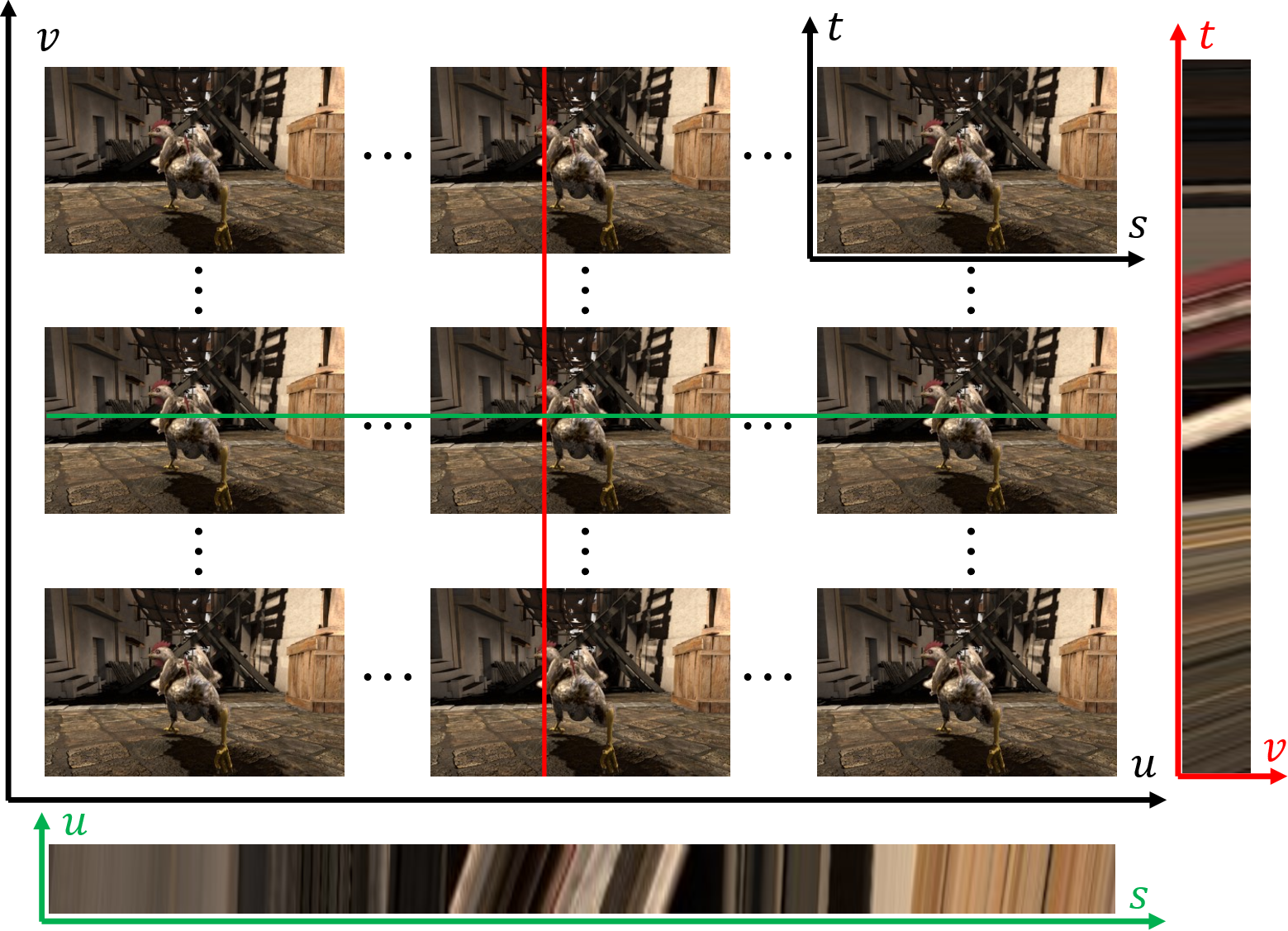}\\
      \subcaption{4D LFI and corresponding horizontal and vertical EPIs.}
      ~\\
      \label{fig:EPIs}
\end{minipage}
   \hfill
   \begin{minipage}{0.49\hsize}
      \centering
      \includegraphics[height=4.5cm]{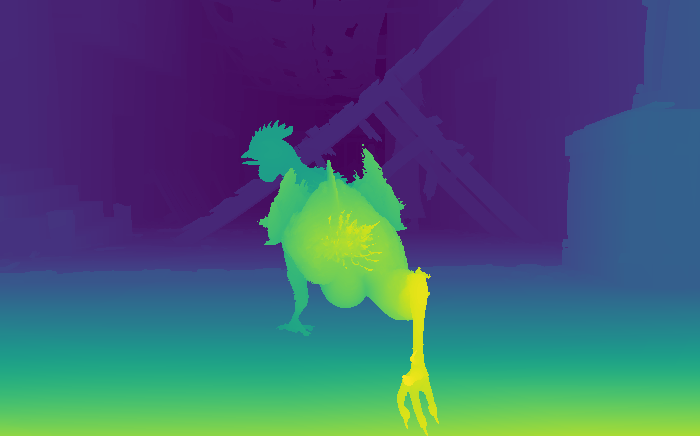}\\
      \subcaption{The ground-truth depth (disparity) map corresponding to central view of 4D LFI.}
      \label{fig:disparity}
   \end{minipage}
   ~\\~\\
   \caption{Exsample of the 4D LFV dataset developed in this study.}
   \label{fig:EPIs_disparity}
\end{figure}

\jtextd{
   4D LFIsを用いた奥行き推定は，ここ数年，活発に研究されている．
   プレノプティックカメラの構造上，サブアパーチャー画像間のベースラインが非常に狭く，
   空間分解能と角度分解能の間にトレードオフがある．
   このことが，
   プレノプティックカメラにより撮影された4D LFIsからの
   奥行き推定を困難にしている．
   %
   近年，角度-空間方向の二次元スライスからなるエピポーラ平面画像（Epipolar Plane Images: EPIs）
   （図~\ref{fig:EPIs}）を用いた深層学習ベースの手法
   ~\cite{shin2018epinet,leistner2019learning,faluvegi20193d}が，
   4D LFIs奥行き推定手法を評価するためのベンチマークとしてよく使用される
   HCI 4D Light Field Benchmark\footnote{https://lightfield-analysis.uni-konstanz.de/}
   ~\cite{honauer2016benchmark}の精度指標でトップランクを獲得している．
   しかし，従来研究では静的な4D LFIsを用いた奥行き推定に焦点を当てており，
   4Dライトフィールド映像（Light Field Videos: LFVs）のような時間情報を
   考慮した奥行き推定は行われていない．
   単眼奥行き推定においては，連続した映像フレーム間の時間的相関と一貫性を考慮することで
   性能が向上することが示されており~\cite{zhang2019exploiting}，
   時間情報が4D LF奥行き推定にも役立つことを示唆している．
}

Depth estimation from 4D LFIs has been a
research topic for the last few years.
Due to plenoptic cameras'
structure, the baseline between sub-aperture images
is very narrow
and there exists a trade-off between the spatial and the angular resolution.
This makes depth estimation from 4D LFIs captured by a plenoptic camera challenging.
%
%
Recently, deep learning-based methods~\cite{shin2018epinet,faluvegi20193d}
using epipolar plane images (EPIs)(Fig.~\ref{fig:EPIs}), 
which consist of two-dimensional angular-spatial slices, 
achieved high performance
in the accuracy metrics of the HCI 4D Light Field
Benchmark\footnote{\url{https://lightfield-analysis.uni-konstanz.de/}}~\cite{honauer2016benchmark},
which is often used as a benchmark for evaluating 
4D LFIs depth estimation methods.
However, previous studies have focused on depth estimation from static
4D LFIs while no studies have considered
the temporal information, i.e., 4D light field videos (LFVs).
The success in monocular depth estimation considering the temporal
correlations and consistency among consecutive video frames
~\cite{zhang2019exploiting}
%
suggests that the temporal information can also help
4D LF-based depth estimation.

\jtextd{
  このため，本研究では，4D LFVsを入力とする奥行き推定ニューラルネットワークを提案する．
  提案手法は，各フレームのEPIsボリュームから空間-角度情報を抽出し，
  Convolutional Long Short-Term Memory（CLSTM）
  を用いて時間情報を集約することでシーンの奥行きマップを推定している．
}

Therefore, this paper proposes a depth estimation neural network using
4D LFV as input.
The proposed method estimates depth maps of scenes 
by extracting the spatial-angular information from the EPIs volume at each frame and 
by collecting the temporal information using Convolutional Long Short-Term 
Memory (CLSTM).

\jtextd{
  本研究の貢献を以下に示す．
  \begin{itemize}
  
  \item オープンソースの映画``Sintel''~\cite{sintel2010}
    を使用し，深層学習モデルを十分に訓練可能な中規模4D LFVsデータセットを作成した．
  
  \item 著者らの知る限り，4D LFVに記録された空間-角度-時間情報を用いた奥行き推定は
    本研究が初となる．
  
  \item 合成映像および実撮影映像を用いた実験により，時系列情報が4D LFベースの
    奥行き推定の性能向上に寄与することを示した．
   
  \end{itemize}
}

The contributions of this study are summarized as follows:
\begin{itemize}

\item This study developed a medium-scale synthetic 4D LFV dataset that 
   can be used for training deep learning-based methods from
  the open-source movie ``Sintel''~\cite{sintel2010}.
  
\item To the best of our knowledge, this is the first study on depth estimation from
  the spatial-angular-temporal information recorded in a 4D LFV.
%
  
\item Experimental results using synthetic and real-world 4D LFVs 
   showed that the temporal information helps to improve
   4D LF-based depth estimation performance.
%

\end{itemize}

\jtextd{
  なお，
  提案手法が推定する値は視差であるが（図~\ref{fig:disparity}），
  視差と奥行きは相互に変換が可能である．
  よって，本論文では，読者の理解を容易にするために，
  視差と奥行きの区別が不要な箇所では，これらの言葉を区別なく用いる．
  視差と奥行きを厳密に区別する必要がある箇所では，これらの言葉を区別して議論を行う．
}

In this study, the target value estimated by the proposed method is
the disparity (Fig.~\ref{fig:disparity}).
Since the disparity and depth can be transformed into each other,
this paper uses these terms indistinguishably where it is not necessary to make
the distinction between them
to facilitate the reader's understanding.
%

\section{Related works}
\subsection{4D LFIs depth estimation methods}
\jtextd{
   近年，深層学習に基づく4D LFIs奥行き推定手法が数多く提案され，大きな成果を上げている．
   これらの手法は，従来の最適化に基づく手法~\cite{tao2013depth,zhang2016robust}
   と比較して，精度の向上と計算時間の短縮を実現している．
   Heberら~\cite{heber2016convolutional}はEPIsに含まれる線の傾きを予測するために
   畳み込みニューラルネットワーク（Convolutional Neural Network: CNN）を学習し，
   高次正則化を用いて推定結果を精緻化する手法を提案した．
   Heberら~\cite{heber2017neural}は，U字型エンコーダデコーダ構造のネットワークを用いて
   LFIsから幾何学的情報を抽出し，低計算コストで高品質な奥行きマップを生成することに成功した．
   Shinら~\cite{shin2018epinet}は，完全畳み込みアーキテクチャを用いて高速かつ正確な奥行き推定を行った．
   この研究では，4つの角度方向（水平，垂直，左斜め，右斜め）に
   スタックした
   サブアパーチャー画像を入力とするマルチストリーム構造を導入している．
   Faluv{\'e}giら~\cite{faluvegi20193d}は同様のアーキテクチャにおいて各ストリームで
   3次元畳み込みを使用することで，2つの角度方向スライス（水平，垂直）のみを使用した場合でも
   同等の性能を持つことを示した．
}

Over the past few years, many deep learning-based depth estimation
methods from 4D LFIs have been proposed with great
success~\cite{heber2016convolutional,heber2017neural,shin2018epinet,faluvegi20193d}.
Such methods improve both accuracy and runtime compared to 
conventional optimization-based methods~\cite{tao2013depth,zhang2016robust}.
Heber et al. proposed a method that predicts the slope of lines in
EPI using convolutional neural network (CNN) combined with by the
high-order regularization~\cite{heber2016convolutional}.
%
They also succeeded in producing high-quality results at low
computational cost by extracting geometric information from a LFI
using a U-shaped encoder-decoder architecture~\cite{heber2017neural} .
%
%
Shin et al.~\cite{shin2018epinet} proposed a fully convolutional neural network 
with fast and accurate performance in depth estimation.
They introduced a multi-stream architecture consisting of  multiple inputs
of sub-aperture image stacks along four angular directions (horizontal,
vertical, left- and right-diagonal).
Faluv{\'e}gi et al.~\cite{faluvegi20193d} showed that,
by applying 3D convolution in each stream, the performance was maintained even when only 
two angular directions (horizontal, vertical) were used instead of the four directions.

\subsection{Monocular depth estimation methods}
\jtextd{
   画像認識におけるCNNの成功以来，深層学習に基づく単眼奥行き推定手法が数多く提案されており，
   映像を入力とすることで奥行き推定に時系列情報を利用する研究も存在する
   \cite{karsch2014depth,zhou2018unsupervised,zhang2019exploiting}．
   Karschら~\cite{karsch2014depth}は，局所的な動作の手がかりとオプティカルフローを用いて，
   時間的に一貫した奥行きマップを生成する手法を提案した．
   Zhouら~\cite{zhou2018unsupervised}は，
   バンドル調整を用いて奥行きとカメラポーズを同時に推定し，超解像ネットワークを
   用いて細部を復元する手法を提案した．
   また，Zhangら~\cite{zhang2019exploiting}は，CLSTM
   を
   使用して連続した映像フレーム間の空間的特徴と時間的相関を捉える映像奥行き推定フレームワークを提案した．
   具体的には，各映像フレームから
   学習済み深層CNNを使用して空間特徴を抽出し，
   CLSTMに入力して時間的相関を集約することで奥行きを推定する．
   この映像奥行き推定フレームワークでは使用する空間特徴抽出ネットワークの種類を問わないため，
   他の映像奥行き推定手法にも最小限の変更で適用可能である．
}

With the success of CNN in image recognition, 
many deep learning-based monocular depth estimation methods have been proposed, 
and some of them estimate depth values from videos by considering 
the temporal information
~\cite{karsch2014depth,zhou2018unsupervised,zhang2019exploiting}.
Karsch et al.~\cite{karsch2014depth} proposed a method
to produce temporally consistent depth maps
using local motion cues and optical flow. 
Zhou et al.~\cite{zhou2018unsupervised} proposed a method that
simultaneously estimates depth and camera pose using bundle adjustment
and recovers the details of the estimated depth map using a
super-resolution network.
Zhang et al.~\cite{zhang2019exploiting} proposed a video depth
estimation framework that captures spatial features as well as
temporal correlations between consecutive video frames using CLSTM.
%
%
In this framework, the trained deep CNN extracts the spatial features
from video frames and the subsequent CLSTM collects temporal
correlations to estimate depth values.
Their framework can handle various kinds
of spatial feature extraction network, and thus can be applied to
other video depth estimation methods with minimum modification.


\section{The proposed method}

\begin{figure}[tbp]
  \centering
  \includegraphics[width=0.8\linewidth]{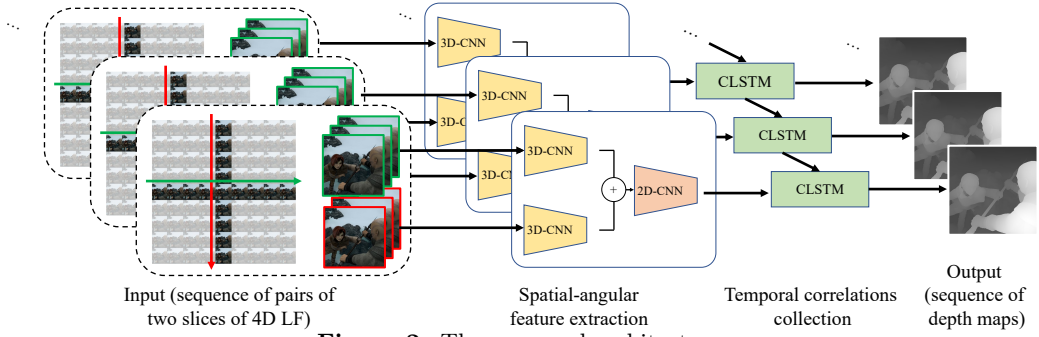}
  \caption{The proposed architecture.}
  \label{fig:architecture}
\end{figure}

\jtextd{
   本研究では，これまでの研究
   ~\cite{zhang2016robust,leistner2019learning,faluvegi20193d}
   の知見に基づき，各4D LFVフレームの2つの角度方向スライス（水平，垂直）のシーケンスを
   入力として使用する奥行き推定ニューラルネットワークアーキテクチャを提案する．
   図\ref{fig:architecture}に示すように，提案アーキテクチャは，
   (1) 2ストリーム3D CNNを用いた2つの角度方向スライスからの空間-角度特徴抽出部と，
   (2) CLSTMを用いた時系列方向の集約による奥行きマップの推定部との2つのモジュールから構成される．
}

This paper proposes a neural network architecture for depth estimation
from 4D LFV, which builds on top of the findings of the previous
studies~\cite{zhang2016robust,faluvegi20193d}.
The proposed architecture uses a temporal sequence of two
image stacks of horizontal and vertical slices of 4D LFV frames as input.
%
Fig.~\ref{fig:architecture} shows the proposed architecture consisting
of two modules:
(1) the first module for the spatial-angular feature extraction from two
LF slices (horizontal and vertical) 
using two-stream 3D CNN, and
(2) the second module for depth estimation 
by collecting the temporal information
of the spatial-angular features
using CLSTM~\cite{zhang2019exploiting}.

\subsection{Two-stream 3D CNN for extracting the spatial-angular features}
\label{sec:two-stream}
\jtextd{
   提案手法では，水平方向と垂直方向の意味のある表現を抽出するために，2ストリーム構造を採用する．
   Faluv{\'e}giら~\cite{faluvegi20193d}
   のモデルにおける
   学習パラメータ数の少なさに着目し，提案手法においても空間-角度特徴抽出のために3D CNNを用いる．
   ここで，EPIsにおける直線の傾きが奥行きを表すことから，
   3次元畳み込みのための第3の空間次元として視点（角度）インデックスを使用する．
   %
   以下では，$9 \times 9$の角度分解能を持つ4D LFを例として説明する．
}

The proposed method adopts a two-stream structure to extract 
meaningful representations of the horizontal and vertical directions
from the sub-aperture images sliced in their direction.
%
%
Focusing on Faluv{e}gi et al.~\cite{faluvegi20193d}'s model, which
benefits from having fewer parameters to be trained, the proposed
method employs 3D CNN for spatial-angular feature extraction.
Here, the viewpoint (angle) index is the third spatial
dimension for the 3D convolution, since the slope of lines in EPIs
(Fig.~\ref{fig:EPIs}) represents depth.
In the following, we describe an example of a 4D LF with an angular
resolution of $9 \times 9$.

\jtextd{
   提案方式を構成する2つのストリームでは，それぞれ2ステップの前処理を行う．
   まず，入力データに対し，バッチ単位でゼロ平均正規化を行う．
   これによりEPIsの線がより強調され，ネットワークの学習が容易になる．
   次に，
   3次元畳み込みが適用された際に特徴マップの空間次元を維持できるよう，
   $4 \times 4$の空間パディングを適用する．
}

Each of the two streams that process the horizontal and vertical
sub-aperture image stacks first applies preprocesses consisting of two
steps.
%
First, zero-mean normalization is applied to the input data in
batches, resulting in better emphasizing the differences in the lines of EPIs
and making it easier for the network to learn.
Next, a spatial padding of $4\times 4$ is applied so that the spatial
dimension of the feature maps can be maintained when applying 3D
convolution.

\jtextd{
   各ストリームは，ストライド幅が1，サイズが$3 \times 3 \times 3$のカーネルを使用した
   3次元畳み込み層を4層連ねた構成をとる．
   出力する特徴マップ数は，最初の畳み込み層は32個，残り3層の畳み込み層は64個である．
   特徴マップのサイズは，4層の3次元畳み込み層によって，
   ビューの数に対応する
   次元は9から1に縮小され，
   空間次元は入力と同等のサイズに維持される．
}

Each stream consists of four consecutive 3D convolutional layers
with kernel size $3\times 3\times 3$ and stride size one.
The number of feature maps to be output is 
32 in the first convolutional layer and 
64 in the remaining three layers.
The four 3D convolutional layers reduce the dimension of feature maps
corresponding to the number of views from nine to one while keeping
the spatial dimension the same size as the input.

\jtextd{
   2つのストリームで抽出された特徴マップを連結することで得られる128個の特徴マップに対し，
   より高度な表現を生成するために5層の2次元畳み込みを適用する．
   2次元畳み込みはすべて$3 \times 3$カーネルを使用し，
   特徴マップサイズを維持するためにゼロパディングを適用する．
   特徴マップの数は，5層の畳み込み層によって
   64，32，32，32，16と徐々に減少する．
}

The two feature map sets obtained from the two streams are
concatenated (totally 128 maps) and input into the next five
2D convolutional layers
%
in order to produce higher level representation.
The proposed method uses 2D convolutional layers with kernel size $3
\times 3$ and zero padding to maintain the feature map size.
The number of feature maps gradually decreases to 64, 32, 32, 32, and
16 as passing through the five convolutional layers.

\jtextd{
   上記の全ての2次元，3次元畳み込み層において，活性化関数としてReLUを適用する．
   また，このプロセスで使用される全てのカーネルは
   Glorotの一様分布
   を用いて初期化される．
}

All of the above 2D and 3D convolutional layers use rectified
linear unit as the activation function, 
and all kernel weights are initialized with Glorot
initialization.

\subsection{CLSTM for collecting the temporal information}
\begin{figure}[tbp]
  \centering
  \includegraphics[width=0.8\linewidth]{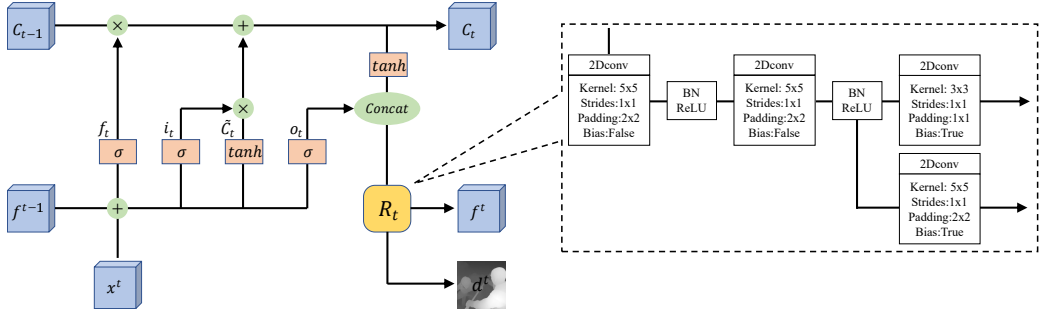}
  \caption{CLSTM structure of the proposed method.}
  \label{fig:CLSTM}
\end{figure}

\jtextd{
   入力フレームは時間次元で連続しているので，これらのフレームの時間的相関を考慮することは
   奥行き推定性能の向上に役立つと考えられる．
   本研究では，Zhangら~\cite{zhang2019exploiting}に倣い，
   CNNによって抽出された空間-角度特徴量の時間情報を集約するために
   図~\ref{fig:CLSTM}に示す構造のCLSTMを用いる．
   CLSTMのセルは次のように表すことができる．
   \begin{equation}
      \begin{aligned}
      f_t &=         \sigma(x^t \ast W_f + f^{t-1} \ast R_f + b_f) \\
      i_t &=         \sigma(x^t \ast W_i + f^{t-1} \ast R_i + b_i) \\
      \tilde{C}_t &= \tanh (x^t \ast W_C + f^{t-1} \ast R_C + b_C) \\
      C_t &=         f_t \otimes C_{t-1} + i_t \otimes \tilde{C}_t \\
      o_t &=         \sigma(x^t \ast W_o + f^{t-1} \ast R_o + b_o) \\
      f^t, d^t &=    R_t([o_t, \tanh(C_t)])
      \end{aligned}
   \end{equation}
   ここで，$\sigma$はシグモイド関数，$\tanh$は双曲線正接関数であり，
   $\otimes$と$\ast$はそれぞれアダマール演算子と畳み込み演算子である．
   $W_f$，$W_i$，$W_C$,
   $W_o$，$R_f$，$R_i$，$R_C$，$R_o$，$b_f$，$b_i$，$b_C$，および，$b_o$は，
   対応する畳み込み層におけるカーネルとバイアス項を表す．
}

As the input frames are continuous in the temporal dimension, 
taking the temporal information of these frames into consideration 
may help to estimate the depth.
Following the previous work~\cite{zhang2019exploiting}, 
the proposed method employs CLSTM shown in Fig.~\ref{fig:CLSTM} 
for collecting the temporal information of the spatial-angular features 
extracted by CNN (Sec.~\ref{sec:two-stream}).
In this study, the number of feature maps in the hidden state $f^{t}$
is set to eight.

\subsection{Loss function}
\jtextd{
  ネットワークの学習のために，Huら~\cite{hu2019revisiting}によって提案された
  損失関数を利用する．
  ただし，Huらは
  奥行きの差異を適切に考慮するために
  奥行きの推定値と真値の差の対数を採用したが，
  本研究では奥行きに反比例する視差を推定するため，
  視差の差異を損失として直接扱う．
  提案手法で用いる損失関数を以下に示す．
  \begin{equation}
    L = l_{depth} + \lambda l_{grad} + \mu l_{normal}
  \end{equation}
  ここで，$\lambda$，$\mu$は重み付け係数である．
  上記の損失関数$L$は3つの項から構成される．
  $l_{depth}$は不正確な奥行き推定に対するペナルティであり
  ，次のように定義する．
  \begin{equation}
    l_{depth} = \frac{1}{n} \sum^{n}_{i=1} \|d_{i} - g_{i}\|_{1}
  \end{equation}
  ここで，$n$は画素数であり，$d_i$と$g_i$はそれぞれ画素$i$の
  推定視差と真値である．
}

This study employs the loss function proposed by Hu et al.~\cite{hu2019revisiting} 
to train the network. 
While Hu et al. adopted the logarithm of 
the difference between the estimated and ground-truth depth value,
this study uses the difference directly as a loss 
since the proposed method estimates disparity that is inversely proportional to depth.
The loss function $L$ consists of three terms, 
designed to penalize inaccurate disparity estimations, the errors around edges,
and small structural errors such as high-frequency undulation of a surface,
respectively:
\begin{equation}
  \begin{aligned}
   L &= l_{disparity} + \lambda l_{grad} + \mu l_{normal} \\
   l_{disparity} &= \frac{1}{n} \sum^{n}_{i=1} \|d_{i} - g_{i}\|_{1} \\
   l_{grad} &= \frac{1}{n} \sum^{n}_{i=1}
    \left(
      \|\nabla_{x}(d_{i}) - \nabla_{x}(g_{i})\|_{1} + 
      \|\nabla_{y}(d_{i}) - \nabla_{y}(g_{i})\|_{1}
    \right)\\
   l_{normal} &= \frac{1}{n} \sum^{n}_{i=1}
     \left(
       1 -
       \frac{\eta^d_i \cdot \eta^g_i}
       {\sqrt{\eta^d_i \cdot \eta^d_i} \sqrt{\eta^g_i \cdot \eta^g_i}}
     \right) \, ,
  \end{aligned}
\end{equation}
where $\lambda$ and $\mu$ are weighting coefficients.
$n$ denotes the number of pixels; 
$d_i$ and $g_i$ are the estimated and ground-truth 
disparity of pixel $i$, respectively.
$\nabla_{x}$ and $\nabla_{y}$ represent 
the spatial derivative along the $x$-axis and $y$-axis respectively.
$\eta^d_i = [-\nabla_x (d_i), -\nabla_y (d_i), 1]$ and
$\cdot$ denotes inner product.

\section{4D Light field video dataset}
\label{sec:dataset}


\jtextd{
   4D LFVsからの奥行き画像生成方式の性能評価を行うために，
   ``Sintel''~\cite{sintel2010}からSintel 4D LFVsデータセットを作成した
   （図~\ref{fig:EPIs_disparity}）．
   既存の公開されている4D LFVsデータセット~\cite{wang2017light,brizzi2019feature}
   では，サンプル数が少ないか，ground-truth奥行きが提供されていないために，
   深層学習ベースの4D LFVs奥行き推定手法の有効性を
   十分に
   検証することが難しい．
   作成したデータセットは，$1,204 \times 436$画素，
   $9 \times 9$視点，
   20～50フレームを持つ23の合成4D LFVで構成され，
   中央ビューにおけるground-truth奥行きを持つため，深層学習モデルを十分に訓練可能である．
}

This study developed the Sintel 4D LFV dataset from the open-source movie ``Sintel''~\cite{sintel2010}
(Fig.~\ref{fig:EPIs_disparity})
because it was difficult to accurately evaluate the effectiveness of 
deep learning-based 4D LFVs depth estimation methods 
with existing 4D LFV datasets~\cite{wang2017light,brizzi2019feature}
due to small number of samples or the lack of ground-truth disparity values.

%
The generated dataset consists of 23 synthetic 4D LFVs 
with $1,204 \times 436$ pixels, $9 \times 9$ views, and 20--50 frames, 
and includes the ground-truth disparity values in the central view, 
so that it can be used for training deep learning-based methods.

\jtextd{
  それぞれのシーンは，MPI Sintelデータセット~\cite{butler2012naturalistic}
  を参考に
  ``Sintel''のプロダクションファイルを修正し，``clean''パスでレンダリングを行った．
  ``clean''パスは，滑らかな陰影や鏡面反射など，複雑な照明と反射特性が含まれている一方で，
  ボケやモーションブラー，半透明オブジェクトは除外されている．
}

Each scene was rendered with a ``clean'' pass 
after modifying the production file of ``Sintel'' 
with reference to the MPI Sintel dataset~\cite{butler2012naturalistic}.
A ``clean'' pass includes
complex illumination and reflectance properties including specular reflections,
such as smooth shading and specular reflections,
while bokeh, motion blur, and semi-transparent objects are excluded.
\jtextd{
  4D LFVsの撮影は，光軸を平行に保ちながらカメラを共通の焦点面に
  向かって移動させて行った．
  視差マップは，Blender
  で得られる奥行きマップを
  次の式を用いて変換することで取得した．
  \begin{equation}
    disparity = \frac{B \times f}{z}
  \end{equation}
  ここで，$B$はベースライン，$f$は焦点距離，$z$は奥行きである．
  この変換により得られた視差マップは，ほとんどのシーンでは[0, 1]の範囲内にあるが，
  いくつかのシーンでは1.5まである．
}
The 4D LFVs were captured by moving the camera
with a baseline of 0.01[m]
towards a common focus plane while keeping the optical axes parallel.
The ground-truth disparity value was obtained by transforming 
the depth value obtained in Blender.
%


\section{Experiments}
\jtextd{
   提案手法の有効性を検証するために，作成したSintel 4D LFVsデータセットお
   よび実撮影4D LFVsを用いて実験を行った．
  奥行き推定に時間情報を考慮しない従来手法~\cite{faluvegi20193d}と比較す
  ることで，時系列情報の利用が奥行き推定の精度向上に寄与するかどうかを検
  討した．
}
The performance of the proposed method is 
evaluated using the Sintel 4D LFV dataset
and the real-world 4D LFVs~\cite{wang2017light}.
By comparing with the conventional method~\cite{faluvegi20193d} as the
baseline that does not consider the temporal information, we examined
whether the temporal information contributes to improve the depth
estimation performance.
The baseline method has a structure that replaces the CLSTM module of
the proposed method with a 2D convolutional layer producing a single
feature map.
%

\subsection{Experimental setup}
\jtextd{
  本実験ではSec.~\ref{sec:dataset}で作成したSintel 4D LFVsデータセットに
  含まれる23シーンを学習用16シーン，検証用3シーン，テスト用4シーンに分割して使用した．
  また，パッチによる学習を行うために，空間分解能が$32 \times 32$，
  フレーム数が5になるように，16の空間ストライド，1の時間ストライドを使用して
  約1.4M個のパッチサンプルを作成した．
  シーン単位の分割により，約1Mサンプルを学習データに割り当て，
  検証データとテストデータにはそれぞれ約130Kサンプル，約300Kサンプルを割り当てた．
}

In the experiments, 23 scenes in the Sintel 4D LFV dataset 
created in Sec.~\ref{sec:dataset} 
were divided into three subsets, i.e., 16 scenes that were used for training, 
three scenes for validation and four scenes for test.
In total, we extracted about 1.4M patch samples with a spatial
resolution of $32 \times 32$ and a frame length of five using a
spatial stride of 16 and a temporal stride of one.
By the scene-wise splitting, about 1M, 130K and 300K samples were
allocated to the training, validation and test data, respectively.

\jtextd{
  提案手法は，ネットワーク全体を包括的に訓練することが可能であり，
  本実験ではepochの上限を20とした．
  Adam optimizer~\cite{kingma2014adam}を用い，バッチサイズを64に設定した．
  初期学習率を0.0005に設定し，10，15epochで0.1倍ずつ減少させることとした．
  また，損失関数の重み付け係数$\lambda$，$\mu$は$\lambda=1$，$\mu=1$に設定した．
  提案手法はTensorFlowをバックエンドとするKerasを使用して実装し，
  学習は1枚のNVIDIA GTX 1080Tiを搭載した計算機上で約4日を要した．
}

Parameters were configured as follows: 
The upper limit of epoch and the batch size were set to 20 and 64, respectively.
Adam optimizer~\cite{kingma2014adam} was employed.
The learning rate started at 0.0005 and was decreased by 
a factor of 0.1 after 10 and 15 epochs.
The weights $\lambda$ and $\mu$ of loss function $L$ were
set as $\lambda=1$ and $\mu=1$.
It took about four days for the proposed network to be trained on a
computer with a NVIDIA GTX 1080Ti.

\jtextd{
  提案手法の評価には，
  4D LFIs奥行き推定で一般的に適用される指標である，
  平均二乗誤差（Mean Square Error: MSE）と不良画素比（BadPix）を使用した~\cite{honauer2016benchmark}．
  BadPixの定義は，絶対誤差が指定された閾値を超える，
  すなわち，$\|d_{i} - g_{i}\|_{1} > t$を満たす画素の比率である．
  ここで$t$は閾値であり，BadPixの計算には3つの閾値（0.07，0.03，0.01）がよく用いられる．
}

The results were measured
using mean square error (MSE) and bad pixel ratio (BadPix),
which are common in 4D LFIs depth estimation~\cite{honauer2016benchmark}.
The definition of BadPix is the percentage of pixels 
whose absolute errors exceed the specified threshold, 
i.e., $\|d_{i} - g_{i}\|_{1} > t$, 
where $t$ is the threshold.
Three thresholds are often used for calculating 
the bad pixel ratios: 0.01, 0.03, and 0.07.
These values were calculated based on $1,024 \times 432$ depth maps that
were reconstructed by combining the depth estimation results for test
patches.

\jtextd{
  本研究における評価では，各テストパッチから奥行きを推定し，空間軸に沿って結合した後，
  $432 \times 1024$の空間分解能を持つ奥行きマップを用いて
  各評価指標を計算したことに注意されたい．
}

\subsection{Results}

\begin{figure}[tbp]
  \begin{center}
    \includegraphics[width=0.9\linewidth]{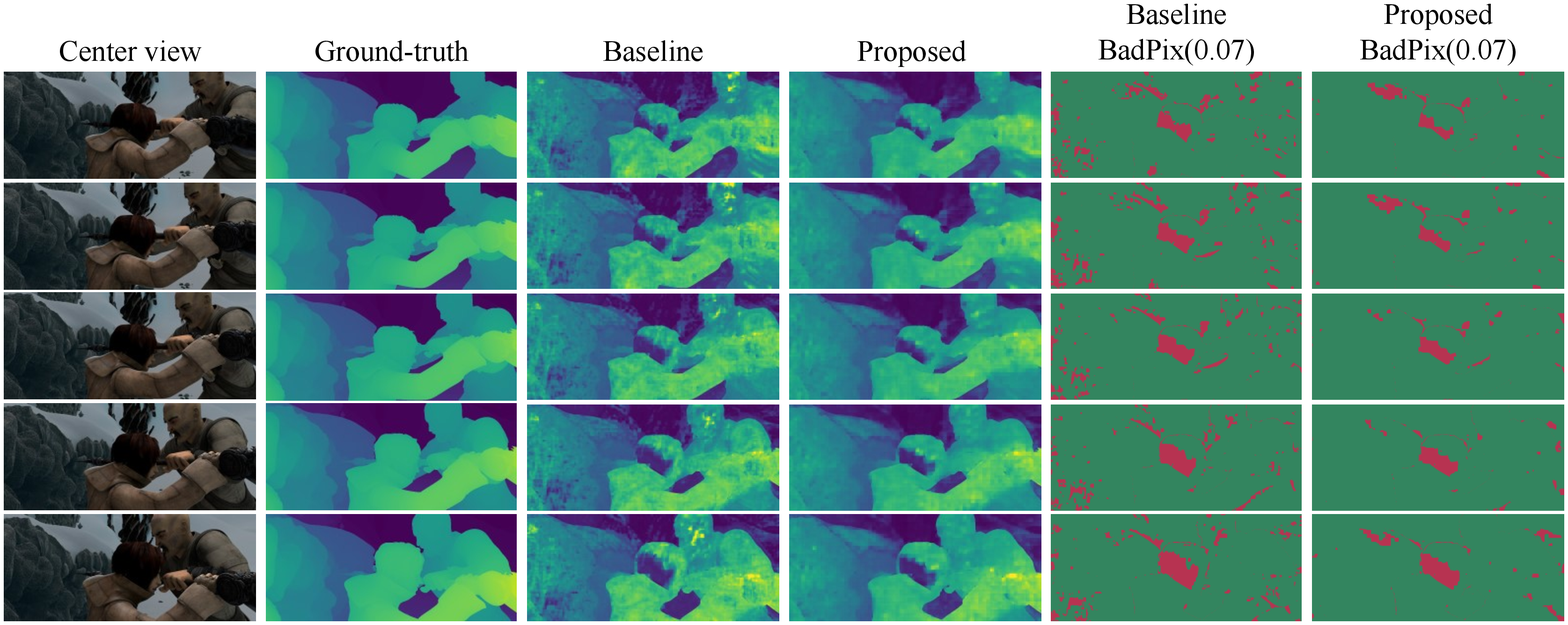}
    \caption{
      The estimation results of the proposed method and the baseline method~\cite{faluvegi20193d}.
    }
    \label{fig:ambushfight5}
  \end{center}
  \begin{minipage}{0.6\textwidth}
    \begin{center}
      \makeatletter
      \def\@captype{table}
      \makeatother
      \caption{
          Comparisons of the propsed method and the baseline method~\cite{faluvegi20193d}
          on the test scenes of the Sintel 4D LFV dataset.
       }
      \scalebox{0.85}{
        \begin{tabular}{l|cccc}
          \hline
          \multicolumn{5}{c}{ambushfight5} \\ \hline
          Method & MSE*100 & BadPix(0.07) & BadPix(0.03) & BadPix(0.01) \\ \hline
          Baseline~\cite{faluvegi20193d} & 0.2590 & 11.005 & 30.3701 & 65.4103 \\
          Proposed & \textbf{0.2167} & \textbf{8.3404} & \textbf{22.8762} & \textbf{62.0493} \\
          \hline
          \multicolumn{5}{c}{bamboo3} \\ \hline
          Method & MSE*100 & BadPix(0.07) & BadPix(0.03) & BadPix(0.01) \\ \hline
          Baseline~\cite{faluvegi20193d} & 0.2895 & 12.1862 & 29.2804 & 57.9944 \\
          Proposed & \textbf{0.2159} & \textbf{8.9475} & \textbf{21.8162} & \textbf{53.2985} \\
          \hline
          \multicolumn{5}{c}{shaman2} \\ \hline
          Method & MSE*100 & BadPix(0.07) & BadPix(0.03) & BadPix(0.01) \\ \hline
          Baseline~\cite{faluvegi20193d} & 3.391 & 37.9157 & 53.4889 & 75.6842 \\
          Proposed & \textbf{2.4421} & \textbf{32.7585} & \textbf{50.6706} & \textbf{74.7733} \\
          \hline
          \multicolumn{5}{c}{thebigfight2} \\ \hline
          Method & MSE*100 & BadPix(0.07) & BadPix(0.03) & BadPix(0.01) \\ \hline
          Baseline~\cite{faluvegi20193d} & \textbf{0.0305} & 1.0964 & 4.4142 & 18.6974 \\
          Proposed & 0.0367 & \textbf{1.0688} & \textbf{3.6084} & \textbf{17.7493} \\
          \hline
        \end{tabular}
      }
      \label{tb:experiment_results}
    \end{center}
  \end{minipage}
  \hfill
  \begin{minipage}{0.4\textwidth}
    \begin{center}
      \includegraphics[width=0.9\linewidth]{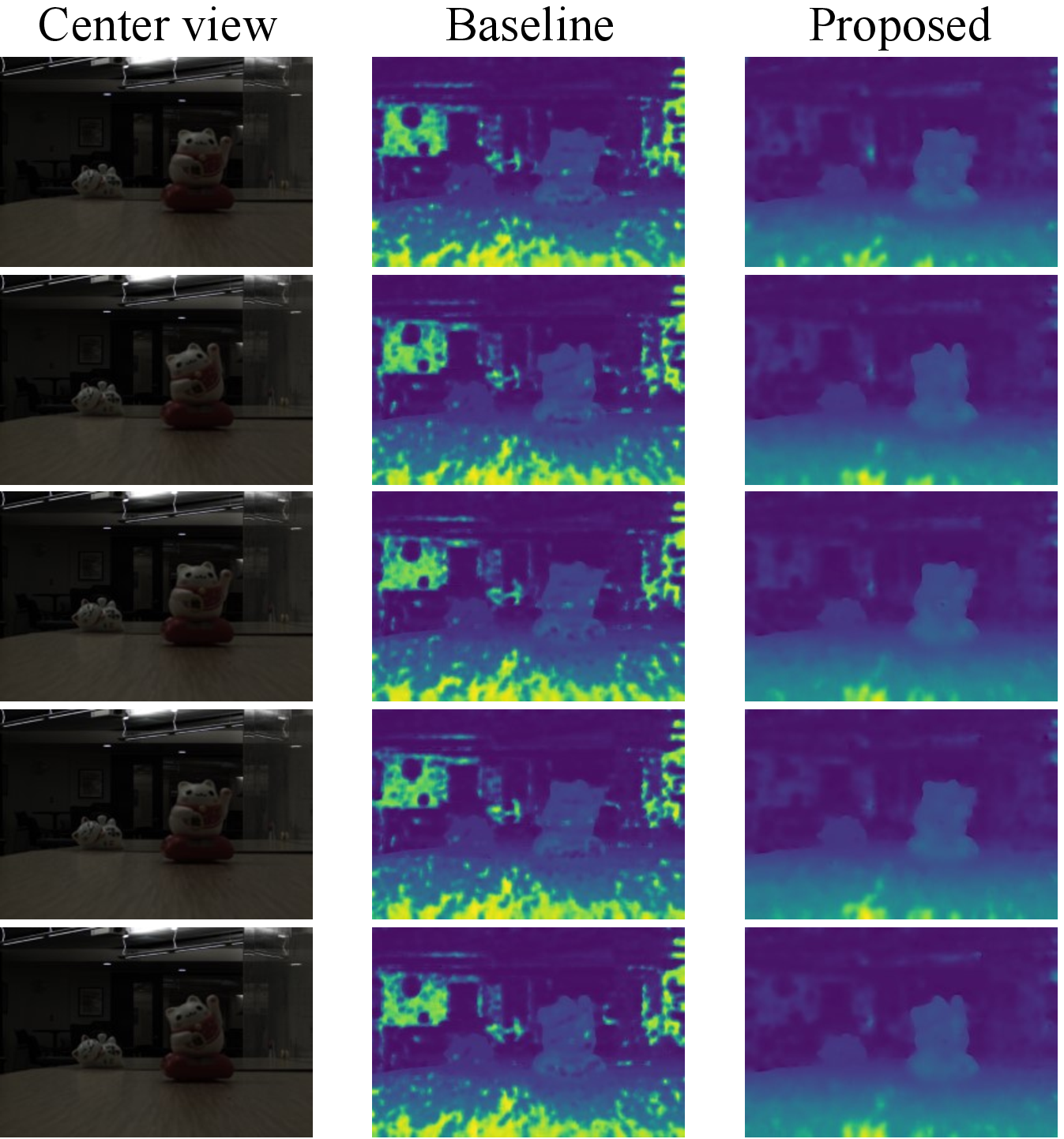}
      \caption{
        The results of real-world data.
      }
      \label{fig:real_results}
    \end{center}
  \end{minipage}
\end{figure}

\jtextd{
  表\ref{tb:experiment_results}に，
  提案手法と時間情報を考慮しない従来手法~\cite{faluvegi20193d}を用いて，
  4つのテストシーンの奥行きを推定した際の評価指標値を示す．
  従来手法は，提案手法のCLSTMモジュールを，
  単一の特徴マップを出力する2次元畳み込み層に置き換えた構造を持つ．
  表\ref{tb:experiment_results}より，
  時間情報を考慮することにより，ほとんどのテストシーンで
  提案手法が優れた奥行き推定性能を発揮することがわかる．
  thebigfight2の結果に着目すると，提案手法は
  MSEではわずかに劣っているもののBadPixでは優れているため，
  従来手法では推定が困難な領域において，
  提案手法はより適切に奥行きを推定できたことがわかる．
}

Table~\ref{tb:experiment_results} shows the performance of 
the proposed method and 
the baseline method~\cite{faluvegi20193d}  
that does not consider the temporal information 
on the four test scenes.
%
%
In most of the test scenes,
the proposed method achieved better performance than the baseline by
considering the temporal information.
Focusing on the results of \textit{thebigfight2}, 
the proposed method was slightly inferior in MSE 
but superior in BadPix, indicating that 
the proposed method estimated better disparity values
in regions that were difficult to estimate for the baseline method.

\jtextd{
  テストシーンambushfight5における推定結果の一部を
  図~\ref{fig:ambushfight5}に示す．
  このシーンは，ノイズの多いテクスチャを持つ岩肌が映り込み，
  2人の登場人物が争う激しい動きを含むシーケンスである．
  この例からもわかるように，提案手法は時間情報を活用することで
  ノイズの多い領域の奥行き推定を可能にする．
  また，従来手法でもエッジをうまく捉えているが，
  提案手法はより表面が滑らかな奥行きマップを推定できている．
}

Fig.~\ref{fig:ambushfight5} shows 
example estimation results for \textit{ambushfight5} scene,
which included noisy textured rock surfaces
and two characters fighting.
As can be seen from Fig.~\ref{fig:ambushfight5}, by using the temporal
information the proposed method can estimate a better disparity map with a
smoother surface in noisy regions than the baseline method.

\jtextd{
  一方，提案手法でも推定が難しい領域（画像中心の後頭部）
  が存在することも確認できる．
  この領域に含まれるパッチは，Shinら~\cite{shin2018epinet}によって定義された
  テクスチャレス領域，すなわち，パッチ内の中心画素と他の画素間の平均絶対差が
  0.02未満の領域であることが確認された．
  時間情報を考慮しても，テクスチャレス領域の奥行き推定は難しいことが
  わかった．
}

On the other hand, some regions were difficult to estimate the depth
even for the proposed method, e.g. the occipital region in the image center.
This region corresponded to textureless regions
defined by Shin et al.~\cite{shin2018epinet},
i.e., the mean absolute difference between a center pixel and 
other pixels in the patch was less than 0.02.
%

\jtextd{
  また，テストシーンshaman2における推定結果の一部を
  図~\ref{fig:shaman2}に示す．
  shaman2はambushfight5よりも視差範囲が大きいシーンである．
  BadPixの減少から従来手法と比較して推定性能が向上していることは
  確認できるが，提案手法でも良好な推定結果が得られているとは言いがたい．
  これは，Leistnerら\cite{leistner2019learning}が指摘しているように，
  学習した視差範囲よりも大きな視差を含むシーンでは奥行き推定が難しいためであり，
  時間情報を考慮しても同様の結果であることを示している．
}
\mycut{
Some of estimation results for the shaman2 scene included 
in the test scenes are shown in Fig.~\ref{fig:shaman2}.
The shaman2 is a scene with a larger disparity range 
than the ambushfight5.
The improvement of BadPix shows that the estimation performance 
of the proposed method is better than the baseline method,
but it is difficult to say that 
the proposed method estimates accurate disparity values.
This is because, 
as pointed by Leistner et al. \cite{leistner2019learning}, 
it is difficult to estimate disparity valuess outside of the trained disparity range.
We show that this problem occurs even considering the temporal information.
}

\jtextd{
  前節のCGレンダリングデータに加えて，
  Wangら~\cite{wang2017light}がLytro illumで撮影した4D LFVsを使用して
  提案手法の性能を検証した．
  ライトフィールドカメラで撮影された画像は，
  カメラに内在する構造上の問題により深刻な画像ノイズが含まれているため，より挑戦的である．
  図~\ref{fig:real_results}に，
  Sintel 4D LFVsデータセットで学習されたモデルを使用し，
  パッチ単位ではなくオリジナルの空間分解能を持つシーケンスを入力したときの
  奥行き推定結果を示す．
  実世界データにおいても提案手法は従来手法よりも
  自然な奥行き推定が行えていることがわかる．
  この結果から，本研究で作成した合成4D LFVsデータセットが
  十分写実的であることも示された．
}

In addition to
the synthetic data, 
this study also tested the proposed method using real-world 4D LFVs~\cite{wang2017light}.
%
The real-world data captured by a light field camera is challenging
as they contain severe image noise due to 
the inherent structural problem in the camera.
Fig.~\ref{fig:real_results} shows the disparity estimation results
using the model trained on the Sintel 4D LFV dataset. 
%
We observed that the proposed method provided more natural 
disparity estimation
than the baseline method.
The results also showed
that the synthetic 4D LFV dataset 
created in this study was sufficiently realistic to train deep neural networks.


\section{Conclusions}
\jtextd{
   4D LFVsから奥行きを推定するためのエンドツーエンドの
   ニューラルネットワークアーキテクチャを提案した．
   提案手法は，2ストリームCNNとCLSTMを使用して，
   4D LFVsに記録された空間-角度-時間情報を処理することができる．
   また，深層学習モデルを十分にトレーニング可能な規模の
   4D LFVsデータセットが存在しないため，
   オープンソースの映画``Sintel''
   を使用し，中規模4D LFVsデータセットを作成した．
   実験結果は，時間情報が4D LFベースの奥行き推定性能を
   向上させることを示しており，
   ノイズの多い実世界データにおいても奥行き推定が
   可能であることを示している．
}

This paper proposes an end-to-end neural network architecture 
for depth estimation from 4D LFV.
The proposed method employs two-stream CNN and CLSTM to process 
the spatial-angular-temporal information recorded in a 4D LFV.
This study also constructs a synthetic 4D LFV dataset 
from the open-source movie ``Sintel''~\cite{sintel2010},
which can be used for training deep learning-based methods. 
Experimental results showed that the temporal information contributed to 
the performance of 4D LF-based depth estimation
and that depth estimation was possible even in noisy real-world data.

\jtextd{
  本手法を改善する余地は多く残っている．
  第一に，学習ベースの手法を改善する簡単な方法は学習データの数を増やすことである．
  4D LFIsのデータ水増し手法~\cite{shin2018epinet}を利用することで，
  より汎化性能を高めたモデルの実現が期待できる．
  第二に，テクスチャレス領域の正確な奥行き推定のために，大きな受容野を使用することである．
  空間-角度特徴抽出のバックボーンアーキテクチャとしてU-Net~\cite{ronneberger2015u}を
  使用することで，
  テクスチャレス領域の推定性能が向上することが示されており\cite{leistner2019learning}，
  本研究の提案アーキテクチャに組み込むことで性能の向上が期待できる．
}
\mycut{
  There are still rooms for improving the proposed method.
First, the easiest way to improve the deep learning-based approach is 
to boost the number of training dataset. 
By using the data augmentation method of 4D LFIs~\cite{shin2018epinet}, 
it is expected to be realized more generalized model.
Second, we can use a large receptive field for 
accurate depth estimation of the textureless regions.
It has been shown that 
using U-Net~\cite{ronneberger2015u} as the backbone architecture 
for spatial-angular feature extraction 
improves the estimation performance of textureless regions~\cite{leistner2019learning}, 
and its incorporation into the proposed architecture 
is expected to improve the performance.
}

\bibliographystyle{spiebib} 

\end{document}